\documentclass{article}

\usepackage{microtype}
\usepackage{graphicx}
\usepackage{subfigure}
\usepackage{booktabs} 

\usepackage{hyperref}

\usepackage[accepted]{icml2024}

\usepackage{amsmath}
\usepackage{amssymb}
\usepackage{mathtools}
\usepackage{amsthm}
\usepackage{multirow}
\usepackage{colortbl}
\definecolor{lightgray}{rgb}{0.9, 0.9, 0.9}
\usepackage{makecell}
\usepackage[capitalize,noabbrev]{cleveref}

\theoremstyle{plain}

\theoremstyle{definition}

\theoremstyle{remark}

\usepackage[textsize=tiny]{todonotes}

\icmltitlerunning{Memory-Space Visual Prompting for Efficient Vision-Language Fine-Tuning}

\begin{document}

\twocolumn[
\icmltitle{Memory-Space Visual Prompting for Efficient Vision-Language Fine-Tuning}




\begin{icmlauthorlist}
\icmlauthor{Shibo Jie}{pku}
\icmlauthor{Yehui Tang}{huawei}
\icmlauthor{Ning Ding}{pku,huawei}
\icmlauthor{Zhi-Hong Deng}{pku,agi}
\icmlauthor{Kai Han}{huawei}
\icmlauthor{Yunhe Wang}{huawei}
\end{icmlauthorlist}

\icmlaffiliation{pku}{School of Intelligence Science and Technology, Peking University}

\icmlaffiliation{huawei}{Huawei Noah’s Ark Lab}
\icmlaffiliation{agi}{National Key Laboratory of General Artificial Intelligence}


\icmlcorrespondingauthor{Yunhe Wang}{yunhe.wang@huawei.com}
\icmlcorrespondingauthor{Zhi-Hong Deng}{zhdeng@pku.edu.cn}
\icmlcorrespondingauthor{Kai Han}{kai.han@huawei.com}

\icmlkeywords{Machine Learning, ICML}

\vskip 0.3in
]



\printAffiliationsAndNotice{}  

\begin{abstract}

Current solutions for efficiently constructing large vision-language (VL) models follow a two-step paradigm: projecting the output of pre-trained vision encoders to the input space of pre-trained language models as visual prompts; and then transferring the models to downstream VL tasks via end-to-end parameter-efficient fine-tuning (PEFT). However, this paradigm still exhibits inefficiency since it significantly increases the input length of the language models. In this paper, in contrast to integrating visual prompts into inputs, we regard visual prompts as additional knowledge that facilitates language models in addressing tasks associated with visual information. Motivated by the finding that Feed-Forward Network (FFN) of language models acts as  ``key-value memory'', we introduce a novel approach termed memory-space visual prompting (MemVP), wherein visual prompts are concatenated with the weights of FFN for visual knowledge injection. Experimental results across various VL tasks and language models reveal that MemVP significantly reduces the training time and inference latency of the fine-tuned VL models and surpasses the performance of previous PEFT methods. Code: \url{https://github.com/JieShibo/MemVP} 
\end{abstract}

\section{Introduction}
Recently, the investigation of pre-trained foundation models has achieved remarkable  success in the fields of both computer vision and natural language processing~\cite{llama,gpt4,tang2024rethinking,clip}, thereby fostering advancements in vision-language (VL) models. It has been found that VL models can be efficiently constructed upon off-the-shelf pre-trained vision encoders and language models~\cite{frozen,flamingo,blip2,llava}. 
The de-facto paradigm to combine them involves projecting the outputs of vision encoders, \emph{i.e.}, image features, to visual prompts within  the input space of the language models via linear projection or resampler. Subsequently, the language models concentrate the visual prompts with the text embedding tokens and process them as a whole.

\begin{figure}[t]
\centering
\includegraphics[width=0.9\columnwidth]{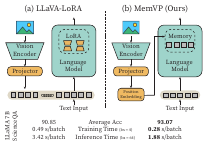}
\vspace{-10pt}
\caption{{Illustration of PEFT methods using (a) the conventional input-space visual prompting  and (b) our memory-space visual prompting.} MemVP outperforms previous paradigms in terms of performance, training speed, and inference speed.}
\vspace{-12pt}
\label{icml-historical}
\end{figure}

Nevertheless, the scale of both vision models and language models is experiencing exponential growth, \emph{e.g.}, ViT-G~\cite{vit-g} has 1.8B parameters and LLaMA~\cite{llama} has up to 70B parameters. Therefore, both pre-training and fine-tuning their combinations with a vast number of parameters for downstream VL tasks become prohibitively expensive in terms of training and storage resources. To mitigate this challenge, parameter-efficient fine-tuning (PEFT) methods incorporate lightweight modules (\emph{e.g.}, adapters~\cite{adapter}, LoRA~\cite{lora}) into the models, and/or select a small subset of pre-trained parameters (\emph{e.g.}, bias, normalization). During fine-tuning, only these modules and selected parameters are updated. Prior studies~\cite{vl-adapter,lavin} have demonstrated that, even without resource-intensive VL pre-training, the combinations of vision encoders and language models can still be transferred to downstream VL tasks via PEFT while matching the performance of full fine-tuning.

Although such ``input-space visual prompting \& PEFT'' paradigm proves efficient for training and storage, its mechanism of visual prompts still limits the inference and training efficiency. For instance, the average length of the text inputs is only 6.3 in VQAv2~\cite{vqav2} dataset and 81 in ScienceQA~\cite{scienceqa} dataset, whereas the number of visual tokens can be up to 256 in LLaVA~\cite{llava}. Consequently, in many scenarios, the input tokens of the language models are mostly visual tokens, thereby significantly amplifying the computation cost during training and inference. 

In this paper, we aim to explore an alternative manner for integrating the visual information into language models for downstream VL tasks, which is intended to not only be parameter-efficient but also facilitate fast training and inference. Existing research~\cite{kvmem} has found that, the Feed-Forward Network (FFN) of language models acts as key-value memory that stores factual association as knowledge, \emph{e.g.}, \emph{``Strawberries are red''} could be such knowledge stored in FFNs. Inspired by this, we infer that the visual information also contains vision-related factual association that is not included in the memory of language models, \emph{e.g.}, the language models do not realize \emph{``The fruits in  the image are red''}. Therefore, it is necessary to inject such external knowledge into language models to enable them to tackle vision-related tasks. Since FFN is the main carrier of knowledge, we can put the visual information in the memory space of language models, \emph{i.e.}, weights of FFN, instead of input space, thus avoiding extending the input length.%

Based on this motivation, we propose Memory-Space Visual Prompting (MemVP), a PEFT framework for adapting pre-trained vision encoders and language models to downstream VL tasks. As shown in Figure~\ref{icml-historical}, MemVP first projects the features extracted by vision encoders to the dimension of language models as visual prompts. The position-embeded visual prompts are concatenated with the weight matrices of the fully-connected (FC) layers in each FFN block of the language models. During fine-tuning, we freeze most parameters of the vision encoders and language models, only the VL projection layers and position embeddings are tunable. Without extending the inputs, MemVP only introduces a very small amount of extra parameters and computation to the language models, and is thus more efficient during training and inference.

To evaluate the efficiency and effectiveness of MemVP, we conduct experiments across various downstream VL benchmarks, including visual question answering on VQAv2, GQA~\cite{gqa}, and ScienceQA, and image captioning on COCO Captions~\cite{caption}. Additionally, we  evaluate MemVP on language models with  different scales and architectures, including BART~\cite{bart} and T5~\cite{t5} with an encoder-decoder architecture, as well as decoder-only LLaMA-7B and LLaMA-13B. MemVP demonstrates superior performance compared to previous PEFT baselines, while achieving remarkable acceleration for both training and inference. 

%

\section{Related Work}
\subsection{Vision-Language Models}
In the field of VL learning, many different model architectures have been proposed to meet the requirements of different VL tasks, such as dual-encoder~\cite{clip}, fusion-encoder~\cite{lxmert,albef,vilt,meter}, encoder-decoder~\cite{vl-t5,simvlm,pali,ofa,blip,blip2,llava}, \emph{etc}. Recently, the rapid advancement of large language models has prompted a growing number of researchers to regard VL tasks as a process of visual-conditioned text generation, and focus on how to involve vision information in off-the-shelf pre-trained language models. For example, BLIP~\cite{blip} and Flamingo~\cite{flamingo} insert new cross-attention layers into the language models to interact with visual features; Frozen~\cite{frozen}, LLaVA~\cite{llava}, and PaLI~\cite{pali} use the vision encoder to generate visual prompts as the inputs of language models. BLIP-2~\cite{blip2} also uses a large Q-former as resampler to reduce the length of visual prompts.

\subsection{Parameter-Efficient Fine-Tuning for VL Alignment}
PEFT has already been widely studied in the field of vision~\cite{adapter-cv,adaptformer,noah,ssf,fact,biadapter}, language~\cite{adapter,adapterp,lora,bitfit,ptuningv2}, and multi-modality~\cite{vl-adapter,vl-pet,lavin,hyperpelt,mixphm,uniadapter}. Particularly, based on the pre-trained vision encoders and language models, the VL models can be trained in a parameter-efficient manner. There are many studies focusing on PEFT of such assembled VL models on downstream tasks. VL-Adapter~\cite{vl-adapter} and VL-PET~\cite{vl-pet} project the image features as visual prompts, and fine-tune the projector and PEFT modules inserted in the T5 or BART models. Differently, LLaMA-Adapter~\cite{llama-adapter} concatenates the visual prompts with the hidden state of LLaMA's intermediate layers. LaVIN~\cite{lavin} inserts adapters in both the vision encoder and LLaMA, and introduces a routing mechanism for adapters.

Through PEFT, it becomes possible to train VL models using off-the-shelf uni-modal models with less time and GPU memory. However, it is noteworthy that these studies do not take computation efficiency into account, which is one of the main contributions of our paper.

\subsection{Memory of Language Models}

\citet{kvmem} discover that the FFN of pre-trained language models is essentially key-value memory which stores factual association. Based on this finding, \citet{ffn1} locate and edit knowledge in language models by replacing certain rows of the matrices of FFN with the embedding of the object. \citet{ffn2} edit the located factual knowledge by adding new key-value pairs to FFN. \citet{ffn3} expand the size of FFN with extra keys and values as a knowledge bank. \citet{ffn4} replace FFN in language models with differentiable plug-in key-value memory for interpretability. However, current works only focus on pure language models, without exploring the potential of visual information as external factual knowledge.

\section{Revisiting Visual Prompts in VL Models}

\begin{figure}[t]
\centering

\includegraphics[width=0.9\columnwidth]{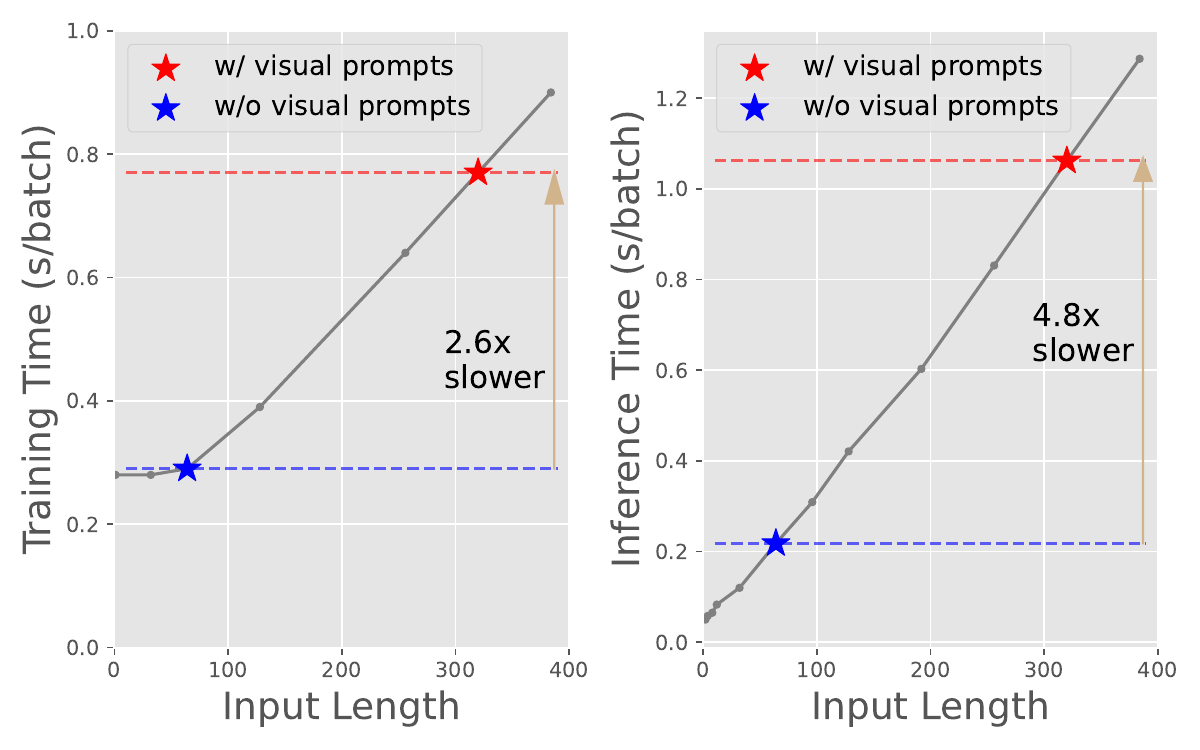}
\vspace{-10pt}
\caption{\textbf{Training and inference time of LLaMA-7B on a single V100.}  The training process adopts PEFT in which we only tune LoRA modules. The training batch size and inference batch size are 4 and 16, respectively, to maximize utilization of GPU memory.  We also highlight the position when the text token length is 64 w/ and w/o input-space visual prompts. The length of visual prompts is 256 as in LLaVA. We fix the output length to 1.}
\label{fig:speed}
\vspace{-10pt}
\end{figure}


Current VL models mostly adopt a common architecture, including a pre-trained vision encoder, a pre-trained language model, and a module that bridges the two components. An efficient bridging module could be one or several FC layers that project the features of the images into the input space of the language model as visual prompts. Although the VL projection of visual prompts is parameter-efficient, it is not computation-efficient enough for training and inference. To obtain fine-grained local visual information, the visual prompts are usually projected from patch features of images, which contain a considerably large number of tokens. For example, LLaVA~\cite{llava} uses ViT-L/14 as vision encoder, which involves 256 tokens to express each image. 
The additional visual prompts significantly increase the length of the input sequence, leading to more computation during training and inference.

\textbf{To what extent do the visual prompts affect the computation speed?} We show the inference speed across different lengths of input and output on LLaMA-7B in Figure \ref{fig:speed}. The computational complexity is $\mathcal{O}(L^2d + Ld^2)$ for  Multi-Head Self-Attention (MHSA) and $\mathcal{O}(LdD)$ for FFN, in which $L$, $d$, and $D$ are the length of token sequence, dimension of tokens, and hidden dimension of FFN, respectively. For example, after applying the visual prompts with 256 tokens to LLaMA-7B as in LLaVA, the training and inference latency of the language model part increase to 2.6$\times$ and 4.8$\times$ on the  text with an input length of 64 and an output length of 1.  

\begin{figure*}[t]
\centering
\includegraphics[width=2.08\columnwidth]{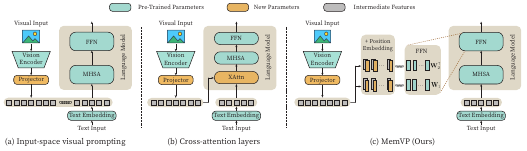}
\vspace{-20pt}
\caption{\textbf{Overview of the mainstream paradigms to concatenate vision encoder and language model.} {\textbf{(a)}} Concatenating visual prompts with the text tokens as inputs of the language model is not computation-efficient, \emph{e.g.}, LLaVA, VL-Adapter, VL-PET. {\textbf{(b)}} Using cross-attention layers to incorporate the visual information from visual tokens is not parameter-efficient, \emph{e.g.}, Flamingo, BLIP. {\textbf{(c)}} Our MemVP injects visual prompts into the FFN blocks of language models, achieving both parameter and computation efficiency.}
\vspace{-5pt}
\label{fig:overview}
\end{figure*}



\textbf{Are there alternative solutions to use fewer visual tokens?} 
BLIP2~\cite{blip2} uses a Q-former as resampler to reduce the number of visual tokens, which compresses the length of visual prompts from 256 to 32. Flamingo~\cite{flamingo} uses a single \texttt{<image>} token as the visual prompt, and insert new resampler and cross-attention to interact with visual features. Although reducing the sequence length, these methods introduce hundreds of millions, or even billions, of new parameters,  which necessitate large-scale VL pre-training. Therefore, we have to perform expensive VL pre-training again when switching to new pre-trained vision encoders or language models. Moreover, since the new modules are large, the training process cannot be parameter-efficient enough to reduce memory and time costs. Also, the large new modules still bring considerably more computation.

Overall, to obtain VL models that are efficient during both training and inference, we need a new paradigm to concatenate pre-trained vision encoders and language models, which \emph{i)} introduces negligible new parameters and extra computation; and \emph{ii)} performs well when PEFT on downstream VL tasks.

\section{Memory-Space Visual Prompting}

\subsection{Preliminary: Reformulation of FFN}

The standard FFN of language models is composed of two FC layers with non-linear activation in-between. Supposing $\boldsymbol{x}\in\mathbb{R}^{d}$ is a input token of the FFN, the FFN can be formulated as:
\begin{equation}
\texttt{FFN}(\boldsymbol{x}) = \phi(\boldsymbol{x} \boldsymbol{W}_1) \boldsymbol{W}_2,    
\end{equation}
in which $\phi$ is activation like ReLU and GELU, $\boldsymbol{W}_1\in\mathbb{R}^{d\times D}$ and $\boldsymbol{W}_2\in\mathbb{R}^{D\times d}$ are the weight matrices of the two FC layers. Note that $\boldsymbol{W}_1$ and  $\boldsymbol{W}_2$ can be rewritten as:
\begin{equation}\boldsymbol{W}_1 = (\boldsymbol{k}_1, \boldsymbol{k}_2, ..., \boldsymbol{k}_D), \boldsymbol{W}_2 = (\boldsymbol{v}_1, \boldsymbol{v}_2, ..., \boldsymbol{v}_D)^\intercal,\end{equation}
in which $\boldsymbol{k}_i\in\mathbb{R}^{d}$ and $\boldsymbol{v}_i\in\mathbb{R}^{d}$ are entries of key and value, respectively. Then, the FFN can be rewritten as
\begin{equation}
    \texttt{FFN}(\boldsymbol{x}) = \sum^D_{i=1}\phi(\langle \boldsymbol{x}, \boldsymbol{k}_i\rangle)\cdot\boldsymbol{v}_i.
    \label{eq:ffn}
\end{equation}
Therefore, the FFN can be interpreted as using input $\boldsymbol{x}$ as the query to calculate its similarity with keys, and gathering values based on the similarity. Previous work has found that FFN acts as a key-value memory storing factual knowledge~\cite{kvmem}.

\subsection{FFN with Visual Prompting}
As illustrated in Figure~\ref{fig:overview}, in conventional input-space visual prompting, the image features are projected to the prefix of the input as context for text generation. Since increasing the input length leads to inefficiency, we avoid using extra visual tokens, and thus all the visual information needs to be contained in textual tokens. A solution to incorporating visual information is to let the textual tokens retrieve information from the visual features. Previous works like Flamingo and BLIP perform retrieval via cross-attention layers, which can be formulated as
\begin{equation}\small\texttt{XAttn}(\boldsymbol{x}) =\texttt{softmax}\left(\frac{\boldsymbol{x}\boldsymbol{W}_q{\boldsymbol{W}_k}^\intercal\boldsymbol{Z}^\intercal}{\sqrt{d}}\right)\boldsymbol{Z}\boldsymbol{W}_v{\boldsymbol{W}_o}^\intercal,\end{equation}
in which $\boldsymbol{x}\in\mathbb{R}^{d}$ is a textual token and $\boldsymbol{Z}=(\boldsymbol{z}_1, \boldsymbol{z}_2,..., \boldsymbol{z}_n)^\intercal\in\mathbb{R}^{n\times d'}$ is the visual features. However, the cross-attention layer introduces a large amount of new parameters, \emph{i.e.}, $\boldsymbol{W}_{q/k/v/o}$, which is far from parameter efficiency and brings considerable additional computation.

Note that the cross-attention  essentially performs a soft look-up using the query $\boldsymbol{x}\boldsymbol{W}_q$ from the key-value pairs $(\boldsymbol{Z}\boldsymbol{W}_k, \boldsymbol{Z}\boldsymbol{W}_v)$ and outputs the weighted average of the retrieved values. Inspired by the fact that FFN also performs similar retrieval from its key-value memory, we consider a more simplified and efficient retrieval process for visual features:
\begin{equation}
    \texttt{Retrieval}(\boldsymbol{x})=\sum_{i=1}^n\phi(\langle\boldsymbol{x}, \mathcal{K}(\boldsymbol{z}_i)\rangle)\cdot \mathcal{V}(\boldsymbol{z}_i),
    \label{eq:attn}
\end{equation}
in which $\mathcal{K}(\boldsymbol{z}_i),\mathcal{V}(\boldsymbol{z}_i)\in\mathbb{R}^{d}$ are the key and value corresponding to $\boldsymbol{z}_i$. This formulation shares a similar form with Eq~(\ref{eq:ffn}). Since the size of FFN's key-value memory $D$ is usually much larger than the number of visual features $n$ ($D=11008$ in LLaMA-7B and $n=256$ for ViT-L/14), the computation of retrieving visual features is insignificant. Therefore, we do not introduce new cross-attention layers as in previous work, but perform such retrieval along with FFN instead. 

From the perspective of FFN, we regard the $\left(\mathcal{K}(\boldsymbol{z}_i),\mathcal{V}(\boldsymbol{z}_i)\right)$ as new memory entries to complement vision-related knowledge that language models used to lack. The new visual key-value entries are inserted into memory,
\begin{equation}
\small\texttt{FFN}(\boldsymbol{x}) =  \sum^D_{i=1}\phi(\langle \boldsymbol{x}, \boldsymbol{k}_i\rangle)\cdot\boldsymbol{v}_i + \sum_{i=1}^n\phi(\langle\boldsymbol{x}, \mathcal{K}(\boldsymbol{z}_i)\rangle)\cdot \mathcal{V}(\boldsymbol{z}_i).
\label{eq3}
\end{equation}
As for $\mathcal{K}$ and $\mathcal{V}$, they should realize two key functions: \emph{i)} aligning the dimension between visual feature $\boldsymbol{z}_i\in\mathbb{R}^{d'}$ and textual token $\boldsymbol{x}\in\mathbb{R}^{d}$, and \emph{ii)} identifying the position of each entry in the visual input. We use a projector $f$, which could be one or several FC layers, to project the visual features to the dimension of the textual token as a visual prompt. The projector is shared between $\mathcal{K}$ and $\mathcal{V}$ for parameter efficiency. The projected visual features are then added with position embedding,
\begin{equation}\mathcal{K}(\boldsymbol{z}_i) = \lambda f(\boldsymbol{z}_i) +  \boldsymbol{p}^{k}_{i}, \quad \mathcal{V}(\boldsymbol{z}_i) = \lambda f(\boldsymbol{z}_i) +  \boldsymbol{p}^{v}_{i},\end{equation}
in which  $\lambda$ is a hyperparameter and $\boldsymbol{p}^{k}, \boldsymbol{p}^{v}\in\mathbb{R}^{n\times d}$ are position embedding for visual prompts inserted into keys and values, respectively. 

To implement Eq~(\ref{eq3}), the position-embedded visual prompts are inserted into the memory as new key-value entries. For the FFN block, the weight matrices are modified to
\begin{equation}\begin{aligned}&\small\boldsymbol{W'}_1 = (\boldsymbol{k}_1, \boldsymbol{k}_2, ..., \boldsymbol{k}_D, \lambda f(\boldsymbol{z}_1) + \boldsymbol{p}^{k}_{1}, ..., \lambda f(\boldsymbol{z}_n) +  \boldsymbol{p}^{k}_{n}),\\
&\small\boldsymbol{W'}_2 = (\boldsymbol{v}_1, \boldsymbol{v}_2, ..., \boldsymbol{v}_D, \lambda f(\boldsymbol{z}_1) + \boldsymbol{p}^{v}_{1}, ..., \lambda f(\boldsymbol{z}_n) + \boldsymbol{p}^{v}_{n})^\intercal.\end{aligned}\end{equation}

Since the visual prompts are concatenated with the FFN weights which are actually memories, we call the proposed new paradigm \emph{memory-space visual prompting} (MemVP).

Besides the standard FFN above, which is widely used in small and middle-scale language models, large language models usually adopt Gated Linear Units (GLU) to enhance the FFN for better performance. For instance, LLaMA uses SwiGLU in FFN, which is
\begin{equation}\texttt{FFN}(\boldsymbol{x}) = (\texttt{SiLU}(\boldsymbol{x} \boldsymbol{W}_1) \otimes \boldsymbol{x} \boldsymbol{W}_{3}) \boldsymbol{W}_2.\label{eq:swiglu}\end{equation}
Supposing $\boldsymbol{W}_{3} = (\boldsymbol{g}_1, ..., \boldsymbol{g}_D)$, Eq~(\ref{eq:swiglu}) can be rewritten as
\begin{equation}
\texttt{FFN}(\boldsymbol{x}) =  \sum^D_{i=1}\texttt{SiLU}(\langle \boldsymbol{x}, \boldsymbol{k}_i\rangle)\cdot\langle \boldsymbol{x}, \boldsymbol{g}_i\rangle\cdot\boldsymbol{v}_i ,
\end{equation}
where $\langle \boldsymbol{x}, \boldsymbol{g}_i\rangle$ can  be viewed as matching the query with another key. 
For FFN using GLU, we simply let the second key entries responding to the visual prompts be $\frac{\boldsymbol{x}}{|\boldsymbol{x}|_2^2}$, \emph{i.e.}, modify $\boldsymbol{W}_{3}$ to
\begin{equation}\boldsymbol{W'}_{3} = (\boldsymbol{g}_1, \boldsymbol{g}_2, ..., \boldsymbol{g}_D, \frac{\boldsymbol{x}}{|\boldsymbol{x}|_2^2}, ..., \frac{\boldsymbol{x}}{|\boldsymbol{x}|_2^2}),\end{equation}
which is equivalent to omitting the second  key when looking up the visual knowledge to avoid involving more parameters, \emph{i.e.},
\begin{equation}
\begin{aligned}
&\texttt{FFN}(\boldsymbol{x}) = \sum^D_{i=1}\texttt{SiLU}(\langle \boldsymbol{x}, \boldsymbol{k}_i\rangle) \cdot\langle \boldsymbol{x}, \boldsymbol{g}_i\rangle\cdot\boldsymbol{v}_i\\
&  +  \sum^n_{i=1}\texttt{SiLU}(\langle \boldsymbol{x}, \lambda f(\boldsymbol{z}_i) + \boldsymbol{p}^{k}_{i}\rangle) \cdot (\lambda f(\boldsymbol{z}_i) + \boldsymbol{p}^{v}_{i}).
\end{aligned}
\end{equation}
In this paradigm, only the projector and position embedding are newly introduced, which are negligible compared with the large size of the pre-trained models. During fine-tuning, we can freeze the parameters of the vision encoders and language models, and only fine-tune these new parameters.

From another perspective, the added key and value entries can be regarded as the two fully connected layers of a vision-conditioned adapter for PEFT. Therefore, in practice, we also adopt some design philosophy of adapters~\cite{lavin}. First, we set the length of position embedding as a hyperparameter to control the number of trainable parameters. We allow the length of position embedding to be longer than the visual prompts, in which case we simply zero-pad the visual prompt to align their lengths. Second, we add another scaling factor to the retrieval results as a hyperparameter to control their magnitude.

\subsection{Complexity Analysis}

We consider a language model layer that is only composed of MHSA and FFN blocks. For simplicity, we omit the bias terms and normalization layers. Let $L$, $d$, and $n$ denote the length of token sequence, dimension of tokens, and length of visual prompts, respectively. The FLOPs of MHSA and FFN are $8Ld^2 + 4L^2d$ and $16Ld^2$ respectively. We use $\texttt{FLOPs}_\textit{LM}$, $\texttt{FLOPs}_\textit{VP}$, and $\texttt{FLOPs}_\textit{MemVP}$ to denote the FLOPs of a single transformer layer in the language model without visual prompts,  with input-space visual prompts, and with memory-space visual prompts, respectively. Then we have
\begin{equation}
    \texttt{FLOPs}_\textit{LM} = 4Ld(6d + L).\end{equation}
For the previous manner which uses input-space visual prompting, the length of the input sequence becomes $L+n$. Then, the additional FLOPs of a layer are
\begin{equation}\texttt{FLOPs}_\textit{VP}-\texttt{FLOPs}_\textit{LM} = 4nd(6d + n + 2L).\end{equation}
Whereas for MemVP, the length of the input is unchanged, and only the hidden dimension of FFN is increased. The additional FLOPs of a layer is
\begin{equation}\texttt{FLOPs}_\textit{MemVP}-\texttt{FLOPs}_\textit{LM} = 4ndL.\end{equation}
Since current VL models basically satisfy $d>>n$, and for VL tasks we have $n > L$ in the most cases, we find that $\texttt{FLOPs}_\textit{VP}$ is multiple times of $\texttt{FLOPs}_\textit{LM}$, but the difference between $\texttt{FLOPs}_\textit{LM}$ and $\texttt{FLOPs}_\textit{MemVP}$ can be ignored. For other architectures such as encoder-decoder model, MemVP mainly reduces the FLOPs of the encoder part. Overall, MemVP is computation-efficient for VL tasks on various language model architectures.

\section{Experiments}
In all the experiments, we follow prior works~\cite{vl-adapter,vl-pet,lavin} adopting a fast and economic adaptation setting, \emph{i.e.}, the resource-intensive VL pre-training stage is not incurred. Although VL pre-training has already been widely used nowadays, our setting has practical significance since it enables low-cost deployment on new foundation models, considering the rapid evolution of language models.

\begin{table*}[t]  
\caption{Results on VQAv2, GQA, and COCO Captions. ``FLOPs'' denotes the average FLOPs \emph{in language models} on test set. We report average performance over three runs on {Karpathy test} split for VQAv2 and COCO Captions, and on {test-dev} split for GQA. All the baseline results are reproduced using the official code of VL-PET~\cite{vl-pet}.}
\centering  
\resizebox{0.85\linewidth}{!}{
\begin{tabular}{lc|cc|cc|cc|c}  
\toprule
\multirow{2}{*}{Method}& \multirow{2}{*}{\makecell[c]{\#Trainable \\Params (M/task)} }&\multicolumn{2}{c|}{VQAv2} & \multicolumn{2}{c|}{GQA} &\multicolumn{2}{c|}{COCO Captions}& \multirow{2}{*}{\makecell[c]{Average\\Score}}  \\ 
&&VQA Score&FLOPs (G)&VQA Score&FLOPs (G)&CIDEr&FLOPs (G)\\
\midrule
\multicolumn{8}{l}{\it BART-base}\\ \midrule
Full Fine-Tuning&141.16&65.4&4.8&53.1&5.3&110.6&6.4&76.4\\
Compacter&3.87&64.2&4.9&52.3&5.4&115.3&6.5&77.3\\
LoRA&3.92&64.8&4.8&52.2&5.3&115.1&6.4&77.4\\
VL-Adapter&3.87&\bf65.5&4.9&53.7&5.4&114.3&6.5&77.8\\
VL-PET&3.84&65.3&5.0&53.9&5.5&\bf120.3&6.6&79.8\\
\rowcolor{lightgray}
MemVP (Ours)&\bf3.78&65.2&\bf1.2&\bf55.1&\bf1.8&120.2&\bf2.8&\bf80.2\\
\midrule
\multicolumn{8}{l}{\it T5-base}\\ \midrule
Full Fine-Tuning&224.54&64.3&9.4&52.0&10.8&112.6&12.9&76.3\\
Compacter&6.11&65.5&9.6&53.6&11.0&113.4&13.2&77.5\\
LoRA&6.05&63.3&9.4&50.8&10.8&113.9&12.9&76.0\\
VL-Adapter&6.10&65.6&9.6&54.4&11.0&113.4&13.2&77.8\\
VL-PET&6.07&65.4&9.8&54.6&11.3&\bf121.2&13.4&80.4\\
\rowcolor{lightgray}
MemVP (Ours)&\bf6.00&\bf65.7&\bf2.3&\bf56.0&\bf3.8&120.8&\bf5.8&\bf80.8\\
\bottomrule
\end{tabular}  
}

\label{tab:bart}  
\end{table*} 
\begin{figure*}
    \centering
\includegraphics[width=1.8\columnwidth]{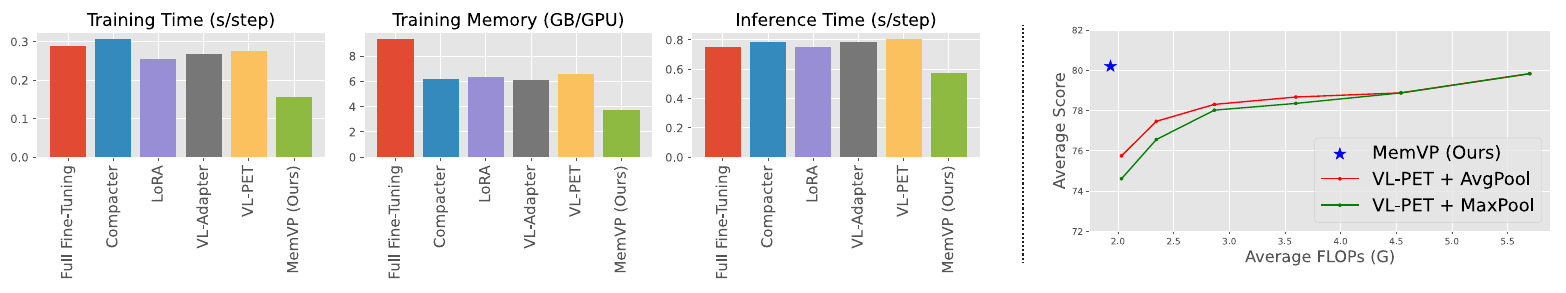}
\vspace{-10pt}
\caption{\textbf{\emph{Left}: Training time, training memory, and inference time of T5-base on VQAv2.} The per-GPU batch sizes for training and inference are 64 and 512, respectively. Measured on V100 GPUs. \textbf{\emph{Right}: Average score  \emph{vs}. FLOPs of BART-base  on the three datasets. } The visual prompts of VL-PET are downsampled to reduce the input length.}
\label{fig:speed2}
\vspace{-5pt}
\end{figure*}

\subsection{Experiments on BART \& T5}

\textbf{Datasets and Baselines. }
For visual question answering, we evaluate our method on VQAv2~\cite{vqav2} and GQA~\cite{gqa}; for image captioning, we evaluate on COCO Captions~\cite{caption}. All these tasks are regraded as text generation tasks which directly output the answers in an open-ended space. Note that, different from previous work~\cite{vl-adapter,vl-pet} using a multi-tasks learning setting where the VQA tasks benefit from the concurrently trained captioning data, we fine-tune MemVP and all the baselines on each dataset individually. We compare MemVP with baselines using previous input-space visual prompting, including current state-of-the-art PEFT methods on BART and T5: VL-Adapter~\cite{vl-adapter} and VL-PET~\cite{vl-pet}, as well as representative PEFT methods designed for language models: Compacter~\cite{compactor} and LoRA~\cite{lora}. We also report the results of fully fine-tuning the language models with input-space visual prompting.

\textbf{Implementation Details. } Following previous work~\cite{vl-adapter,vl-pet}, we use ResNet-101 pre-trained via CLIP~\cite{clip} to pre-extract  image features. The resolution of input images is $224\times224$. The visual encoder is frozen during fine-tuning, and the PEFT modules are only inserted into the language model.  For the language part, we use BART-base~\cite{bart} and T5-base~\cite{t5} with encoder-decoder architecture.  For our MemVP, the grid features before global average pooling are projected to visual prompts via a single FC layer, and the visual prompts are only injected into the FFN blocks of language encoders. Additionally, we also unfreeze the layer normalization of language models. We train on each dataset for 20 epochs with batch size $8\times64$ and report performance on the test set. The hyperparameters of all methods are summarized in Appendix.

\textbf{Results and Analyses. }As shown in Table~\ref{tab:bart}, our MemVP achieves average performance better than current state-of-the-art PEFT method, VL-PET, and much better than other baselines. However, the FLOPs in the language models of MemVP are only $23\%$--$44\%$ of other baselines. To exhibit the advantage of shorter inputs, we compare the training speed, training memory, and inference speed of all methods on VQAv2 in Figure~\ref{fig:speed2} (left).  Compared with VL-PET, MemVP is $1.7\times$ faster during training and $1.4\times$ faster during inference, while using only $56\%$ training memory. Although PEFT only unlocks a small number of parameters for training, the gradient still needs to be propagated back through the whole language model, leading to considerable time and memory consumption. Therefore, the time and memory costs during training and inference are profoundly  affected by the FLOPs in language models.  MemVP releases the time and memory burden for fine-tuning by directly reducing FLOPs, suggesting that computation efficiency is also crucial in designing PEFT methods.

Furthermore, we compared MemVP with a straightforward strategy to reduce the length of the visual prompt: 2D adaptive pooling. As illustrated in Figure~\ref{fig:speed2} (right), after pooling the visual prompt, the input-space prompting methods suffer from obvious performance degradation, implying that the fine-grained local information is lost in this process. By contrast, MemVP can use long visual prompts without extending the input, thus outperforming the baselines in terms of efficiency.

\begin{figure}[t]
\centering
\includegraphics[width=0.9\columnwidth]{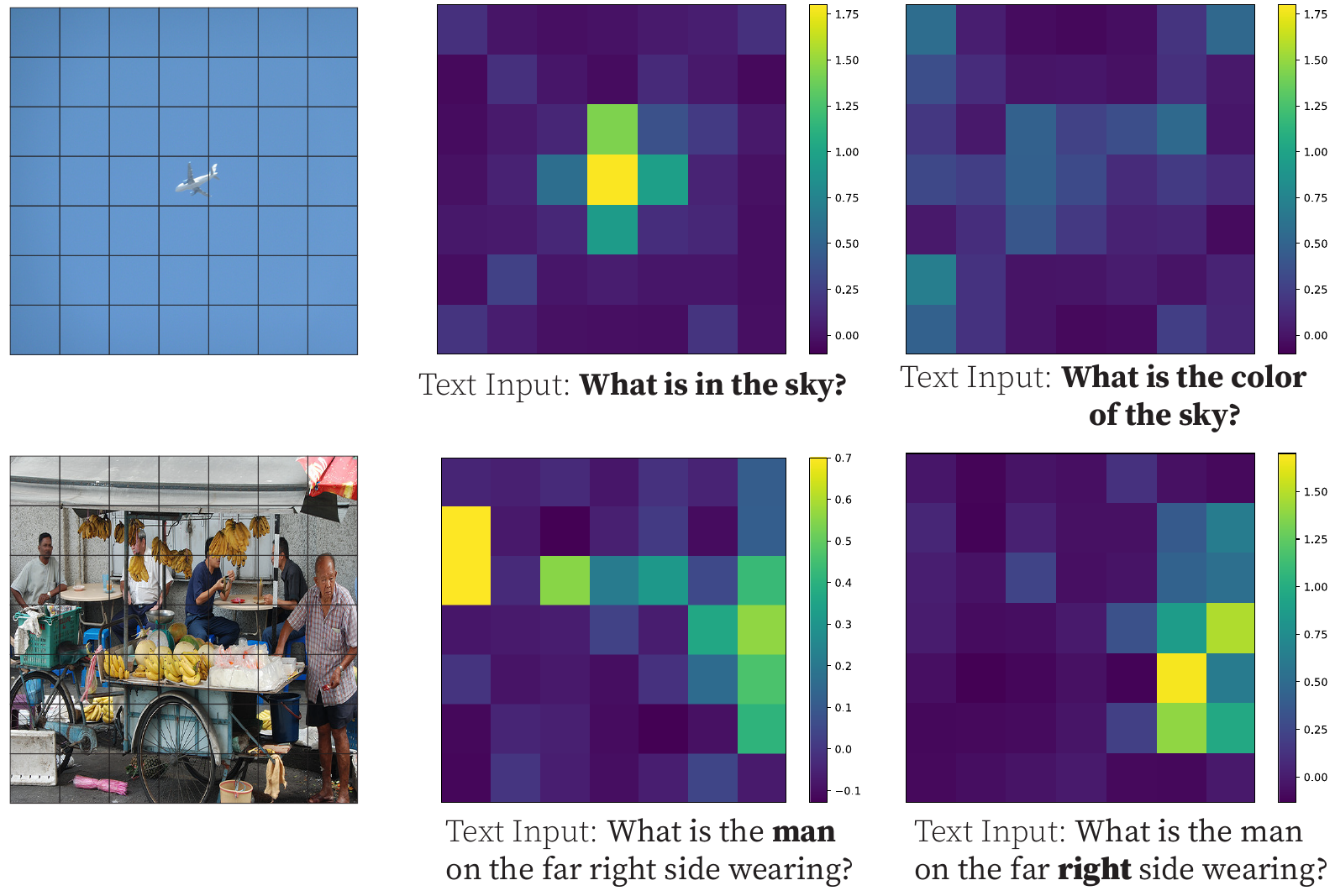}
\vspace{-10pt}
\caption{{\bf Visual knowledge locating.} The similarity values between {\bf blod} text tokens and keys of visual knowledge  are averaged over all layers.}
\vspace{-10pt}
\label{fig:loc}
\end{figure}

\begin{figure}[t]
\centering
\includegraphics[width=0.9\columnwidth]{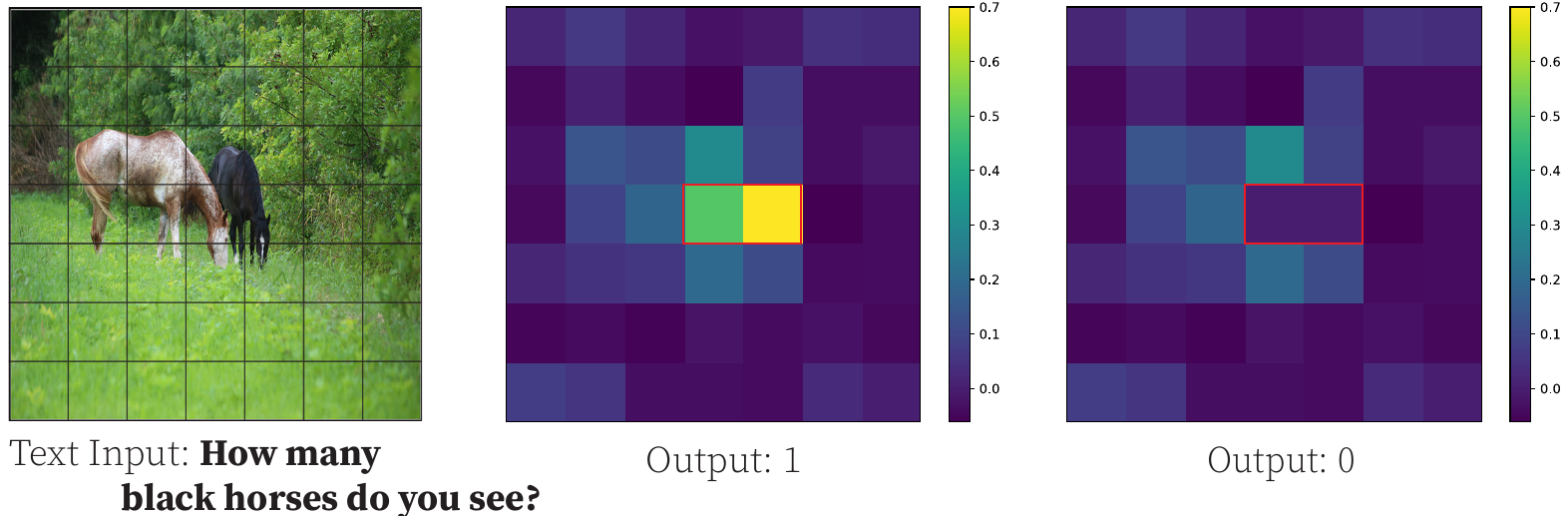}
\vspace{-10pt}
\caption{{\bf Visual knowledge distortion.} \textbf{\emph{Left:}} Inputs of model; \textbf{\emph{Middle:}} Original similarity  between text tokens and keys of visual knowledge; \textbf{\emph{Right:}} Distorted similarity. The values in the red rectangle are set to 0.}
\label{fig:dis}
\vspace{-10pt}
\end{figure}

\textbf{Visualization. }We conduct experiments to verify our main motivation, \emph{i.e.}, the visual information can be inserted into memories of language models as external knowledge. If the model acquires the visual knowledge successfully, we are supposed to observe that \emph{i)} the visual knowledge related to the text inputs is retrieved, and  \emph{ii)} when the model fails to retrieve the correct knowledge under manual distortion, the model should output the corresponding wrong contents.

In Figure~\ref{fig:loc}, we visualize the similarity between queries and keys, \emph{i.e.}, $\phi(\langle \boldsymbol{x}, \lambda f(\boldsymbol{z}_i)\rangle)$ in Eq~(\ref{eq3}), of BART-base fine-tuned on VQAv2. We find that the text tokens have a high similarity with the keys of related visual knowledge entries, implying that the corresponding values are retrieved. For instance, when asking the model \emph{``What is in the sky?''}, the model retrieves knowledge entries around the plain; when asked \emph{``What is the color of the sky?''}, the model retrieves knowledge entries of the background. Moreover, we find that different words in the input sentence have different preferences, \emph{e.g.}, when asking the model \emph{``What is the man on the far right side wearing?''}, the \emph{``man''} token retrieves the knowledge entries that contain men, and the \emph{``right''} token retrieves the entries on the right side of the image.

Next, we try distorting the knowledge by editing the query-key similarity. As the example in Figure~\ref{fig:dis}, when asking the model \emph{``How many black horses do you see?''}, the model mainly retrieves the entries containing the black horse. Then, we manually block the retrieval of the two most responsive entries by setting $\phi(\langle \boldsymbol{x}, \lambda f(\boldsymbol{z}_i)\rangle) = 0$. As a result, the model outputs \emph{``0''} since it fails to obtain knowledge about the existence of black horse. Overall, these observations verify that the visual information is actually inserted into memory and direct the outputs of language models.

\subsection{Experiments on LLaMA}

\begin{table*}[t]  
\caption{ \textbf{Accuracy on ScienceQA test set.}  Question categories: NAT = natural science, SOC = social science, LAN = language
science, TXT = w/ text context, IMG = w/ image context, NO = no context, G1-6 = grades 1-6, G7-12 = grades 7-12. $^\dagger$ denotes our reproduced results. Other results are quoted from their original papers.} 
\centering  
\resizebox{0.9\linewidth}{!}{
\begin{tabular}{lccc|ccc|ccc|cc|c}  
\toprule
\multirow{2}{*}{Method} &\multirow{2}{*}{\makecell[c]{\#Trainable\\ Params}} &\multirow{2}{*}{\makecell[c]{Language\\Model}} &\multirow{2}{*}{\makecell[c]{VL \\Pre-Train}}   & \multicolumn{3}{c|}{Subject} & \multicolumn{3}{c|}{Context Modality} & \multicolumn{2}{c|}{Grade} &  \multirow{2}{*}{Average} \\
     &&&& NAT   & SOC   & LAN   & TXT   & IMG   & NO    & G1-6  & G7-12 &    \\ 
    \midrule
 Human &-& -& -& 90.23 & 84.97 & 87.48 & 89.60 & 87.50 & 88.10 & 91.59 & 82.42 & 88.40 \\
 GPT-4 (0-shot) &-& GPT-4 &-& 84.06 & 73.45 & 87.36 & 81.87 & 70.75 & 90.73 & 84.69 & 79.10 & 82.69 \\ 

LLaVA &7B &  Vicuna-7B  & $\surd$&- & - & - & - & - & - & - & - & 89.84 \\ 
LLaVA &13B &  Vicuna-13B  & $\times$&- & - & - & - & - & - & - & - & 85.81 \\ 

LLaVA &13B &   Vicuna-13B  & $\surd$& 90.36 & {\bf95.95} & 88.00 & 89.49 & 88.00 & 90.66 & 90.93 & {\bf90.90} & 90.92 \\ 

\midrule
\multicolumn{2}{l}{\it PEFT methods}&&&&&&&&&&\\ 

LLaMA-Adapter &1.8M & LLaMA-7B &$\times$& 84.37 &  88.30  &  84.36  & 83.72 &  80.32 &  86.90 &  85.83 &  84.05 & 85.19 \\
LLaVA-LoRA$^\dagger$  &4.4M & LLaMA-7B &$\times$& 91.70 &  94.60 & 86.09  & 91.25 &  90.28 &  88.64 &  91.52 & 89.65 & 90.85 \\

LaVIN &3.8M &  LLaMA-7B&$\times$& 89.25& 94.94& 85.24& 88.51& 87.46& 88.08& 90.16& 88.07& 89.41\\
\rowcolor{lightgray}
MemVP (Ours) &3.9M &  LLaMA-7B&$\times$& 94.45 &{95.05} &	88.64&	93.99	&92.36	&90.94&	93.10&	93.01 &93.07\\
LaVIN &5.4M &  LLaMA-13B&$\times$& 90.32& 94.38 &87.73 &89.44& 87.65& 90.31& 91.19& 89.26& 90.50\\
\rowcolor{lightgray}
MemVP (Ours) &5.5M &  LLaMA-13B&$\times$& {\bf95.07}&\bf95.15&{\bf90.00}&{\bf94.43}&{\bf92.86}&{\bf92.47}&{\bf93.61}&{\bf94.07}&{\bf93.78}\\


\bottomrule
\end{tabular}  
 }
 
\vspace{-5pt}
\label{tab:sqa}  
\end{table*}

\begin{table}[t]  
\setlength{\tabcolsep}{1.5pt}
\caption{\textbf{Training and inference time.} Measured on 8$\times$A800 GPUs without memory-saving or speed-up techniques (\emph{e.g.,} flash attention). The per-GPU batch size is 4 for training and 64 for inference.} 
\centering  
\resizebox{1.0\linewidth}{!}{
\begin{tabular}{lcccc}  
\toprule
Method&\makecell[c]{Length of\\Visual Prompt}&\makecell[c]{\#Trainable\\ Params}&\makecell[c]{Training \\Time (s/batch)}&\makecell[c]{Inference \\Time (s/batch)}\\\midrule
LLaVA-LoRA \small{\ 7B}&256&4.4M&0.49&3.42\\
LaVIN \small{\ 7B} &6&3.8M&0.39&2.06\\
\rowcolor{lightgray}MemVP \small{\ 7B}&256&3.9M&\bf0.28&\bf1.88\\
\rowcolor{lightgray}MemVP \small{\ 13B}&256&5.5M&0.46&3.07\\
\bottomrule
\end{tabular}  
}

\label{tab:speed}  
\vspace{-10pt}
\end{table}

\begin{table}[t]  
\caption{\textbf{Ablation experiments on ScienceQA.} ``Average'' and ``IMG'' denote the accuracy on the whole test set and on the IMG subset, respectively.} 
\setlength{\tabcolsep}{3pt}
\centering  
\resizebox{1.0\linewidth}{!}{
\begin{tabular}{lccc}  
\toprule
Settings&Average&IMG&\makecell[c]{\#Trainable\\ Params (M)}\\\midrule
\rowcolor{lightgray}MemVP \small{7B}&\bf93.07&\bf92.36&3.9\\
\quad w/o visual prompts&85.33&76.05&3.3\\
\quad visual features: local $ \rightarrow$ global &89.01&84.18&3.9\\
\quad position embedding: add $ \rightarrow$ concat&89.79&86.07&3.9\\
\quad insert visual prompts in keys only&91.94&90.23&3.9\\
\quad insert visual prompts in values only&92.78&\bf92.36&3.9\\
\bottomrule
\end{tabular}  
}

\label{tab:ab}  
\vspace{-5pt}
\end{table} 

\textbf{Datasets and Baselines. }
We use a challenging VQA task, ScienceQA~\cite{scienceqa}, to evaluate our method. ScienceQA is a large-scale science question-answering dataset compiled from diverse  knowledge domains. We compare MemVP with other LLaMA-based fine-tuned models with input-space visual prompting, including LLaVA~\cite{llava}, LLaMA-Adapter~\cite{llama-adapter}, and LaVIN~\cite{lavin}. We also provide results of LLaVA equipped with LoRA. All these methods adopt a one-stage paradigm, \emph{i.e.}, directly generating the answers end-to-end without multi-stage chain-of-thought (CoT) prompting~\cite{mm-cot}. We adopt the training recipe used by \citet{lavin} and train each method for 20 epochs. All these methods use a ViT-L/14 pre-trained via CLIP as the visual encoder. We also report zero-shot results of GPT4~\cite{gpt4}.

\textbf{Implementation Details. }
Following LLaVA~\cite{llava}, MemVP and LLaVA-LoRA use the 256 patch features before the last layer of ViT-L/14 and project them as visual prompts. Differently, LaVIN and LLaMA-adapter stack 6 global features (\emph{i.e.}, \texttt{[CLS]} tokens of ViT) selected from different intermediate layers as much shorter visual prompts. The projectors of MemVP, LaVIN, and LLaVA-LoRA are two FC layers with non-linear activation in between.  Since LaVIN also inserts adapters in the visual encoder,  we adopt a comparable strategy on MemVP and LLaVA-LoRA for a fair comparison. Specifically, we introduce parallel adapters to the FFN of the vision encoder following previous work~\cite{adaptformer}. Moreover, since LLaMA has much more layers and larger dimension than BART and T5, we also share the position embedding of MemVP across different layers for parameter efficiency. For the samples that do not have image inputs, we simply set  the visual prompts of MemVP to zero tensors, and only insert the position embedding. 

\textbf{Results and Analyses. }As shown in Table~\ref{tab:sqa}, 
our MemVP significantly outperforms all the baseline PEFT methods on both LLaMA-7B and LLaMA-13B. LLaVA-LoRA performs better than LaVIN and LLaMA-Adapter, indicating that VL models benefit from the local visual information in longer visual prompts. Notably, MemVP also beats LLaVA, a fully fine-tuned model with VL pre-training, on average results as well as 7 out of 8 subsets. Besides, we also compare the training and inference speed of different PEFT methods in Table~\ref{tab:speed}. In spite of the long visual prompts, MemVP is still $1.4\times$ faster than LaVIN during training, since the routing mechanism of LaVIN delays the training speed. LLaVA-LoRA, which also uses local visual prompts in input space, is $1.75\times$ and $1.8\times$ slower than MemVP in training and inference, respectively. Overall, memory-space prompting exhibits remarkable advantage in computation efficiency.

To demonstrate the effectiveness of the components of MemVP, we conduct comprehensive ablation experiments. As in Table~\ref{tab:ab}, when we insert the position embedding without adding visual prompts into the language model, its performance on IMG subset degrades significantly, since the language model cannot obtain the visual knowledge. We note that using global features as in LaVIN leads to a drop in performance due to the loss of local information. We also attempt to concatenate the position embedding with visual prompts instead of adding to them, where the visual prompts will not acquire hard-coded position information but the number of trainable parameters keeps unchanged. The degraded performance indicates the importance of position information for visual prompts since the text inputs may be location-related.  When only inserting visual prompts in keys or values, the model performs worse in both cases.

\section{Conclusion \& Limitation}
In this paper, we revisit the current mainstream ``input-space visual prompting \& PEFT'' paradigm for efficiently bridging pre-trained vision encoders and language models, and point out its remaining inefficiency in terms of computation due to the extended inputs. Motivated by previous findings that the FFNs of language models serve as knowledge memories, we propose MemVP, a memory-space visual prompting method that inserts visual prompts into the FFN weights. Experiments on both small-scale and large-scale language models indicate that MemVP is both parameter-efficient and computation-efficient. Compared to previous state-of-the-art PEFT methods, it achieves competitive performance while enabling faster training and inference, and less memory overhead, providing an economic solution for model adaptation on downstream VL tasks.

However, MemVP still has limitations. Its main contribution lies in reducing the length of the input sequence, but the impact of input length on inference speed primarily occurs during the prefill stage, \emph{i.e.}, the generation of the first token. Therefore, for the generation of long texts (\emph{e.g.}, detailed captioning), MemVP's advantage in inference speed will be diminished, which could be improved in future work.

\section*{Impact Statement}

The method proposed in this paper is based on pre-trained models, especially large language models (LLMs). It may inherit the drawbacks of LLMs such as inherent biases and misinformation, or generate copyright-violating material such as verbatim snippets from non-free content.

\bibliography{main.bib}

\begin{thebibliography}{53}
\providecommand{\natexlab}[1]{#1}
\providecommand{\url}[1]{\texttt{#1}}
\expandafter\ifx\csname urlstyle\endcsname\relax
  \providecommand{\doi}[1]{doi: #1}\else
  \providecommand{\doi}{doi: \begingroup \urlstyle{rm}\Url}\fi

\bibitem[Alayrac et~al.(2022)Alayrac, Donahue, Luc, Miech, Barr, Hasson, Lenc, Mensch, Millican, Reynolds, Ring, Rutherford, Cabi, Han, Gong, Samangooei, Monteiro, Menick, Borgeaud, Brock, Nematzadeh, Sharifzadeh, Binkowski, Barreira, Vinyals, Zisserman, and Simonyan]{flamingo}
Alayrac, J., Donahue, J., Luc, P., Miech, A., Barr, I., Hasson, Y., Lenc, K., Mensch, A., Millican, K., Reynolds, M., Ring, R., Rutherford, E., Cabi, S., Han, T., Gong, Z., Samangooei, S., Monteiro, M., Menick, J.~L., Borgeaud, S., Brock, A., Nematzadeh, A., Sharifzadeh, S., Binkowski, M., Barreira, R., Vinyals, O., Zisserman, A., and Simonyan, K.
\newblock Flamingo: a visual language model for few-shot learning.
\newblock In \emph{Proceedings of NeurIPS}, 2022.

\bibitem[Chen et~al.(2022)Chen, Ge, Tong, Wang, Song, Wang, and Luo]{adaptformer}
Chen, S., Ge, C., Tong, Z., Wang, J., Song, Y., Wang, J., and Luo, P.
\newblock Adaptformer: Adapting vision transformers for scalable visual recognition.
\newblock In \emph{Proceedings of NeurIPS}, 2022.

\bibitem[Chen et~al.(2015)Chen, Fang, Lin, Vedantam, Gupta, Doll{\'{a}}r, and Zitnick]{caption}
Chen, X., Fang, H., Lin, T., Vedantam, R., Gupta, S., Doll{\'{a}}r, P., and Zitnick, C.~L.
\newblock Microsoft {COCO} captions: Data collection and evaluation server.
\newblock \emph{arXiv preprint}, arXiv:1504.00325, 2015.

\bibitem[Chen et~al.(2023)Chen, Wang, Changpinyo, Piergiovanni, Padlewski, Salz, Goodman, Grycner, Mustafa, Beyer, Kolesnikov, Puigcerver, Ding, Rong, Akbari, Mishra, Xue, Thapliyal, Bradbury, and Kuo]{pali}
Chen, X., Wang, X., Changpinyo, S., Piergiovanni, A.~J., Padlewski, P., Salz, D., Goodman, S., Grycner, A., Mustafa, B., Beyer, L., Kolesnikov, A., Puigcerver, J., Ding, N., Rong, K., Akbari, H., Mishra, G., Xue, L., Thapliyal, A.~V., Bradbury, J., and Kuo, W.
\newblock Pali: {A} jointly-scaled multilingual language-image model.
\newblock In \emph{Proceedings of ICLR}, 2023.

\bibitem[Cheng et~al.(2023)Cheng, Lin, Chen, Zhao, and Yan]{ffn4}
Cheng, X., Lin, Y., Chen, X., Zhao, D., and Yan, R.
\newblock Decouple knowledge from paramters for plug-and-play language modeling.
\newblock In \emph{Findings of ACL}, 2023.

\bibitem[Cho et~al.(2021)Cho, Lei, Tan, and Bansal]{vl-t5}
Cho, J., Lei, J., Tan, H., and Bansal, M.
\newblock Unifying vision-and-language tasks via text generation.
\newblock In \emph{Proceedings of ICML}, 2021.

\bibitem[Chu et~al.(2023)Chu, Qiao, Lin, Xu, Yang, Hu, Wei, Zhang, Zhang, Wei, and Shen]{mobilevlm}
Chu, X., Qiao, L., Lin, X., Xu, S., Yang, Y., Hu, Y., Wei, F., Zhang, X., Zhang, B., Wei, X., and Shen, C.
\newblock Mobilevlm : {A} fast, strong and open vision language assistant for mobile devices.
\newblock \emph{arXiv preprint}, arXiv:2312.16886, 2023.

\bibitem[Dai et~al.(2022)Dai, Dong, Hao, Sui, Chang, and Wei]{ffn1}
Dai, D., Dong, L., Hao, Y., Sui, Z., Chang, B., and Wei, F.
\newblock Knowledge neurons in pretrained transformers.
\newblock In \emph{Proceedings of ACL}, 2022.

\bibitem[Dai et~al.(2023)Dai, Jiang, Dong, Lyu, and Sui]{ffn3}
Dai, D., Jiang, W., Dong, Q., Lyu, Y., and Sui, Z.
\newblock Neural knowledge bank for pretrained transformers.
\newblock In \emph{Proceedings of NLPCC}, 2023.

\bibitem[Dou et~al.(2022)Dou, Xu, Gan, Wang, Wang, Wang, Zhu, Zhang, Yuan, Peng, Liu, and Zeng]{meter}
Dou, Z., Xu, Y., Gan, Z., Wang, J., Wang, S., Wang, L., Zhu, C., Zhang, P., Yuan, L., Peng, N., Liu, Z., and Zeng, M.
\newblock An empirical study of training end-to-end vision-and-language transformers.
\newblock In \emph{Proceedings of CVPR}, 2022.

\bibitem[Geva et~al.(2021)Geva, Schuster, Berant, and Levy]{kvmem}
Geva, M., Schuster, R., Berant, J., and Levy, O.
\newblock Transformer feed-forward layers are key-value memories.
\newblock In \emph{Proceedings of EMNLP}, 2021.

\bibitem[Goyal et~al.(2017)Goyal, Khot, Summers{-}Stay, Batra, and Parikh]{vqav2}
Goyal, Y., Khot, T., Summers{-}Stay, D., Batra, D., and Parikh, D.
\newblock Making the {V} in {VQA} matter: Elevating the role of image understanding in visual question answering.
\newblock In \emph{Proceedings of CVPR}, 2017.

\bibitem[Houlsby et~al.(2019)Houlsby, Giurgiu, Jastrzebski, Morrone, de~Laroussilhe, Gesmundo, Attariyan, and Gelly]{adapter}
Houlsby, N., Giurgiu, A., Jastrzebski, S., Morrone, B., de~Laroussilhe, Q., Gesmundo, A., Attariyan, M., and Gelly, S.
\newblock Parameter-efficient transfer learning for {NLP}.
\newblock In \emph{Proceedings of ICML}, 2019.

\bibitem[Hu et~al.(2022)Hu, yelong shen, Wallis, Allen-Zhu, Li, Wang, Wang, and Chen]{lora}
Hu, E.~J., yelong shen, Wallis, P., Allen-Zhu, Z., Li, Y., Wang, S., Wang, L., and Chen, W.
\newblock Lo{RA}: Low-rank adaptation of large language models.
\newblock In \emph{Proceedings of ICLR}, 2022.

\bibitem[Hu et~al.(2023)Hu, Li, Lyu, and Wang]{vl-pet}
Hu, Z., Li, Y., Lyu, M.~R., and Wang, L.
\newblock {VL-PET:} vision-and-language parameter-efficient tuning via granularity control.
\newblock In \emph{Proceedings of ICCV}, 2023.

\bibitem[Hudson \& Manning(2019)Hudson and Manning]{gqa}
Hudson, D.~A. and Manning, C.~D.
\newblock {GQA:} {A} new dataset for real-world visual reasoning and compositional question answering.
\newblock In \emph{Proceedings of CVPR}, 2019.

\bibitem[Jiang \& Zheng(2023)Jiang and Zheng]{mixphm}
Jiang, J. and Zheng, N.
\newblock Mixphm: Redundancy-aware parameter-efficient tuning for low-resource visual question answering.
\newblock In \emph{Proceedings of CVPR}, 2023.

\bibitem[Jie \& Deng(2023)Jie and Deng]{fact}
Jie, S. and Deng, Z.-H.
\newblock Fact: Factor-tuning for lightweight adaptation on vision transformer.
\newblock In \emph{Proceedings of AAAI}, 2023.

\bibitem[Jie et~al.(2023)Jie, Wang, and Deng]{biadapter}
Jie, S., Wang, H., and Deng, Z.
\newblock Revisiting the parameter efficiency of adapters from the perspective of precision redundancy.
\newblock In \emph{Proceedings of ICCV}, 2023.

\bibitem[Kim et~al.(2021)Kim, Son, and Kim]{vilt}
Kim, W., Son, B., and Kim, I.
\newblock Vilt: Vision-and-language transformer without convolution or region supervision.
\newblock In \emph{Proceedings of ICML}, 2021.

\bibitem[Lei et~al.(2018)Lei, Yu, Bansal, and Berg]{tvqa}
Lei, J., Yu, L., Bansal, M., and Berg, T.~L.
\newblock {TVQA:} localized, compositional video question answering.
\newblock In \emph{Proceedings of EMNLP}, 2018.

\bibitem[Lewis et~al.(2020)Lewis, Liu, Goyal, Ghazvininejad, Mohamed, Levy, Stoyanov, and Zettlemoyer]{bart}
Lewis, M., Liu, Y., Goyal, N., Ghazvininejad, M., Mohamed, A., Levy, O., Stoyanov, V., and Zettlemoyer, L.
\newblock {BART:} denoising sequence-to-sequence pre-training for natural language generation, translation, and comprehension.
\newblock In \emph{Proceedings of ACL}, 2020.

\bibitem[Li et~al.(2021)Li, Selvaraju, Gotmare, Joty, Xiong, and Hoi]{albef}
Li, J., Selvaraju, R.~R., Gotmare, A., Joty, S.~R., Xiong, C., and Hoi, S.~C.
\newblock Align before fuse: Vision and language representation learning with momentum distillation.
\newblock In \emph{Proceedings of NeurIPS}, 2021.

\bibitem[Li et~al.(2022)Li, Li, Xiong, and Hoi]{blip}
Li, J., Li, D., Xiong, C., and Hoi, S. C.~H.
\newblock {BLIP:} bootstrapping language-image pre-training for unified vision-language understanding and generation.
\newblock In \emph{Proceedings of ICML}, 2022.

\bibitem[Li et~al.(2023)Li, Li, Savarese, and Hoi]{blip2}
Li, J., Li, D., Savarese, S., and Hoi, S. C.~H.
\newblock {BLIP-2:} bootstrapping language-image pre-training with frozen image encoders and large language models.
\newblock In \emph{Proceedings of ICML}, 2023.

\bibitem[Li et~al.(2020)Li, Chen, Cheng, Gan, Yu, and Liu]{how2qa}
Li, L., Chen, Y., Cheng, Y., Gan, Z., Yu, L., and Liu, J.
\newblock {HERO:} hierarchical encoder for video+language omni-representation pre-training.
\newblock In \emph{Proceedings of EMNLP}, 2020.

\bibitem[Lian et~al.(2022)Lian, Zhou, Feng, and Wang]{ssf}
Lian, D., Zhou, D., Feng, J., and Wang, X.
\newblock Scaling {\&} shifting your features: {A} new baseline for efficient model tuning.
\newblock In \emph{Proceedings of NeurIPS}, 2022.

\bibitem[Liu et~al.(2023{\natexlab{a}})Liu, Li, Li, and Lee]{llava1.5}
Liu, H., Li, C., Li, Y., and Lee, Y.~J.
\newblock Improved baselines with visual instruction tuning.
\newblock \emph{arXiv preprint}, arXiv:2310.03744, 2023{\natexlab{a}}.

\bibitem[Liu et~al.(2023{\natexlab{b}})Liu, Li, Wu, and Lee]{llava}
Liu, H., Li, C., Wu, Q., and Lee, Y.~J.
\newblock Visual instruction tuning.
\newblock In \emph{Proceedings of NeurIPS}, 2023{\natexlab{b}}.

\bibitem[Liu et~al.(2021)Liu, Ji, Fu, Du, Yang, and Tang]{ptuningv2}
Liu, X., Ji, K., Fu, Y., Du, Z., Yang, Z., and Tang, J.
\newblock P-tuning v2: Prompt tuning can be comparable to fine-tuning universally across scales and tasks.
\newblock \emph{arXiv preprint}, arXiv:2110.07602, 2021.

\bibitem[Lu et~al.(2023)Lu, Ding, Huo, Yang, Lu, Tomizuka, and Zhan]{uniadapter}
Lu, H., Ding, M., Huo, Y., Yang, G., Lu, Z., Tomizuka, M., and Zhan, W.
\newblock Uniadapter: Unified parameter-efficient transfer learning for cross-modal modeling.
\newblock \emph{arXiv preprint}, arXiv:2302.06605, 2023.

\bibitem[Lu et~al.(2022)Lu, Mishra, Xia, Qiu, Chang, Zhu, Tafjord, Clark, and Kalyan]{scienceqa}
Lu, P., Mishra, S., Xia, T., Qiu, L., Chang, K., Zhu, S., Tafjord, O., Clark, P., and Kalyan, A.
\newblock Learn to explain: Multimodal reasoning via thought chains for science question answering.
\newblock In \emph{Proceedings of NeurIPS}, 2022.

\bibitem[Luo et~al.(2023)Luo, Zhou, Ren, Chen, Sun, and Ji]{lavin}
Luo, G., Zhou, Y., Ren, T., Chen, S., Sun, X., and Ji, R.
\newblock Cheap and quick: Efficient vision-language instruction tuning for large language models.
\newblock In \emph{Proceedings of NeurIPS}, 2023.

\bibitem[Mahabadi et~al.(2021)Mahabadi, Henderson, and Ruder]{compactor}
Mahabadi, R.~K., Henderson, J., and Ruder, S.
\newblock Compacter: Efficient low-rank hypercomplex adapter layers.
\newblock In \emph{Proceedings of NeurIPS}, 2021.

\bibitem[Meng et~al.(2022)Meng, Bau, Andonian, and Belinkov]{ffn2}
Meng, K., Bau, D., Andonian, A., and Belinkov, Y.
\newblock Locating and editing factual knowledge in {GPT}.
\newblock \emph{arXiv preprint}, arXiv:2202.05262, 2022.

\bibitem[OpenAI(2023)]{gpt4}
OpenAI.
\newblock {GPT-4} technical report.
\newblock \emph{arXiv preprint}, arXiv:2303.08774, 2023.

\bibitem[Pfeiffer et~al.(2021)Pfeiffer, Kamath, R{\"{u}}ckl{\'{e}}, Cho, and Gurevych]{adapterp}
Pfeiffer, J., Kamath, A., R{\"{u}}ckl{\'{e}}, A., Cho, K., and Gurevych, I.
\newblock Adapterfusion: Non-destructive task composition for transfer learning.
\newblock In \emph{Proceedings of EACL}, 2021.

\bibitem[Radford et~al.(2021)Radford, Kim, Hallacy, Ramesh, Goh, Agarwal, Sastry, Askell, Mishkin, Clark, Krueger, and Sutskever]{clip}
Radford, A., Kim, J.~W., Hallacy, C., Ramesh, A., Goh, G., Agarwal, S., Sastry, G., Askell, A., Mishkin, P., Clark, J., Krueger, G., and Sutskever, I.
\newblock Learning transferable visual models from natural language supervision.
\newblock In \emph{Proceedings of ICML}, 2021.

\bibitem[Raffel et~al.(2020)Raffel, Shazeer, Roberts, Lee, Narang, Matena, Zhou, Li, and Liu]{t5}
Raffel, C., Shazeer, N., Roberts, A., Lee, K., Narang, S., Matena, M., Zhou, Y., Li, W., and Liu, P.~J.
\newblock Exploring the limits of transfer learning with a unified text-to-text transformer.
\newblock \emph{J. Mach. Learn. Res.}, 21:\penalty0 140:1--140:67, 2020.

\bibitem[Rebuffi et~al.(2017)Rebuffi, Bilen, and Vedaldi]{adapter-cv}
Rebuffi, S., Bilen, H., and Vedaldi, A.
\newblock Learning multiple visual domains with residual adapters.
\newblock In \emph{Proceedings of NIPS}, 2017.

\bibitem[Sung et~al.(2022)Sung, Cho, and Bansal]{vl-adapter}
Sung, Y., Cho, J., and Bansal, M.
\newblock {VL-ADAPTER:} parameter-efficient transfer learning for vision-and-language tasks.
\newblock In \emph{Proceedings of CVPR}, 2022.

\bibitem[Tan \& Bansal(2019)Tan and Bansal]{lxmert}
Tan, H. and Bansal, M.
\newblock {LXMERT:} learning cross-modality encoder representations from transformers.
\newblock In \emph{Proceedings of the EMNLP-IJCNLP}, 2019.

\bibitem[Tang et~al.(2024)Tang, Liu, Ni, Tian, Bai, Hu, Liu, Jui, Han, and Wang]{tang2024rethinking}
Tang, Y., Liu, F., Ni, Y., Tian, Y., Bai, Z., Hu, Y.-Q., Liu, S., Jui, S., Han, K., and Wang, Y.
\newblock Rethinking optimization and architecture for tiny language models.
\newblock \emph{arXiv preprint}, arXiv:2402.02791, 2024.

\bibitem[Touvron et~al.(2023)Touvron, Lavril, Izacard, Martinet, Lachaux, Lacroix, Rozi{\`{e}}re, Goyal, Hambro, Azhar, Rodriguez, Joulin, Grave, and Lample]{llama}
Touvron, H., Lavril, T., Izacard, G., Martinet, X., Lachaux, M., Lacroix, T., Rozi{\`{e}}re, B., Goyal, N., Hambro, E., Azhar, F., Rodriguez, A., Joulin, A., Grave, E., and Lample, G.
\newblock Llama: Open and efficient foundation language models.
\newblock \emph{arXiv preprint}, arXiv:2302.13971, 2023.

\bibitem[Tsimpoukelli et~al.(2021)Tsimpoukelli, Menick, Cabi, Eslami, Vinyals, and Hill]{frozen}
Tsimpoukelli, M., Menick, J., Cabi, S., Eslami, S. M.~A., Vinyals, O., and Hill, F.
\newblock Multimodal few-shot learning with frozen language models.
\newblock In \emph{Proceedings of NeurIPS}, 2021.

\bibitem[Wang et~al.(2022{\natexlab{a}})Wang, Yang, Men, Lin, Bai, Li, Ma, Zhou, Zhou, and Yang]{ofa}
Wang, P., Yang, A., Men, R., Lin, J., Bai, S., Li, Z., Ma, J., Zhou, C., Zhou, J., and Yang, H.
\newblock {OFA:} unifying architectures, tasks, and modalities through a simple sequence-to-sequence learning framework.
\newblock In \emph{Proceedings of ICML}, 2022{\natexlab{a}}.

\bibitem[Wang et~al.(2022{\natexlab{b}})Wang, Yu, Yu, Dai, Tsvetkov, and Cao]{simvlm}
Wang, Z., Yu, J., Yu, A.~W., Dai, Z., Tsvetkov, Y., and Cao, Y.
\newblock Simvlm: Simple visual language model pretraining with weak supervision.
\newblock In \emph{Proceedings of ICLR}, 2022{\natexlab{b}}.

\bibitem[Zaken et~al.(2022)Zaken, Goldberg, and Ravfogel]{bitfit}
Zaken, E.~B., Goldberg, Y., and Ravfogel, S.
\newblock Bitfit: Simple parameter-efficient fine-tuning for transformer-based masked language-models.
\newblock In \emph{Proceedings of ACL}, 2022.

\bibitem[Zhai et~al.(2022)Zhai, Kolesnikov, Houlsby, and Beyer]{vit-g}
Zhai, X., Kolesnikov, A., Houlsby, N., and Beyer, L.
\newblock Scaling vision transformers.
\newblock In \emph{Proceedings of CVPR}, 2022.

\bibitem[Zhang et~al.(2023{\natexlab{a}})Zhang, Han, Zhou, Hu, Yan, Lu, Li, Gao, and Qiao]{llama-adapter}
Zhang, R., Han, J., Zhou, A., Hu, X., Yan, S., Lu, P., Li, H., Gao, P., and Qiao, Y.
\newblock Llama-adapter: Efficient fine-tuning of language models with zero-init attention.
\newblock \emph{arXiv preprint}, arXiv:2303.16199, 2023{\natexlab{a}}.

\bibitem[Zhang et~al.(2022)Zhang, Zhou, and Liu]{noah}
Zhang, Y., Zhou, K., and Liu, Z.
\newblock Neural prompt search.
\newblock \emph{arXiv preprint}, arXiv:2206.04673, 2022.

\bibitem[Zhang et~al.(2023{\natexlab{b}})Zhang, Guo, Meng, Wang, Wang, Jiang, Liu, and Yang]{hyperpelt}
Zhang, Z., Guo, W., Meng, X., Wang, Y., Wang, Y., Jiang, X., Liu, Q., and Yang, Z.
\newblock Hyperpelt: Unified parameter-efficient language model tuning for both language and vision-and-language tasks.
\newblock In \emph{Findings of ACL}, 2023{\natexlab{b}}.

\bibitem[Zhang et~al.(2023{\natexlab{c}})Zhang, Zhang, Li, Zhao, Karypis, and Smola]{mm-cot}
Zhang, Z., Zhang, A., Li, M., Zhao, H., Karypis, G., and Smola, A.
\newblock Multimodal chain-of-thought reasoning in language models.
\newblock \emph{arXiv preprint}, arXiv:2302.00923, 2023{\natexlab{c}}.

\end{thebibliography}
\bibliographystyle{icml2024}

\newpage
\appendix
\onecolumn
\section*{Appendix}

\section{Experiment Details}
\subsection{Experiments on BART \& T5}
We list the hyperparameters in Table~\ref{tab:hyper-bart}. We use task-specific prompt to the input sentence for each downstream task, as shown in Table~\ref{tab:prompt}.
\begin{table}[h]  
\caption{\textbf{Hyperparameters on BART-base and T5-base.}} 
\centering  
\resizebox{0.95\linewidth}{!}{
\begin{tabular}{lcccc}  
\toprule
Method&Learning Rate&Batch Size&Epoch&Structure Hyper-Parameters\\\midrule
Full Fine-Tuning&1e-4&512&20&-\\
Compacter&1e-3&512&20&hidden dimension $d = 48$, kronecker products $k = 2$\\
LoRA&1e-3&512&20&rank $r = 40$\\
VL-Adapter&1e-3&512&20&hidden dimension $d = 48$\\
VL-PET&1e-3&512&20&\makecell[c]{Encoder: hidden dimension $d = 48$, scaling factor $s = 1$, \#head $N_h= 4$\\ Decoder: hidden dimension $d = 48$, scaling factor $s = 1$, \#head $N_h= 1$}\\
\rowcolor{lightgray}MemVP &1e-3&512&20&scaling factor $\lambda = 0.01$, length of position embedding $n = 240$\\
\bottomrule
\end{tabular}  
}

\label{tab:hyper-bart}  
\end{table}

\begin{table}[h]  
\caption{\textbf{Input-output formats with task prompts.}} 
\centering  
\resizebox{0.95\linewidth}{!}{
\begin{tabular}{lll}  
\toprule
Task&Input&Output\\\midrule
VQAv2&\texttt{[Question]}&\texttt{[Answer]}\\
GQA&\texttt{[Question]}&\texttt{[Answer]}\\
COCO Captions&Provide a one-sentence caption for the provided
image.&\texttt{[Caption]}\\
ScienceQA&Question: \texttt{[Question]}$\backslash$n Context: \texttt{[Context]}$\backslash$n Options:  \texttt{[Choices]}$\backslash$n Reponse: The answer is&\texttt{[Answer]}\\
\bottomrule
\end{tabular}  
}

\label{tab:prompt}  
\end{table} 

\subsection{Experiments on LLaMA}
As mentioned in the main text, since LaVIN also inserts adapters in the visual encoder, we adopt a comparable strategy on MemVP and LLaVA-LoRA for a fair comparison. Specifically, we use two FC layers with hidden dimension of 12 and GELU activation in between as adapters, which are inserted in parallel with the FFNs of ViT, \emph{i.e.}, the adapter use the same input as FFN, and its output is also added to the output of FFN. We list the hyperparameters in Table~\ref{tab:hyper-llama} and the input-output format in Table~\ref{tab:prompt}. 

\begin{table}[h]  
\caption{\textbf{Hyperparameters on LLaMA.}} 
\centering  
\resizebox{1\linewidth}{!}{
\begin{tabular}{lcccc}  
\toprule
Method&Learning Rate&Batch Size&Epoch&Structure Hyper-Parameters\\\midrule
LLaVA-LoRA (7B)&9e-3&32&20& rank $r = 6$\\
\rowcolor{lightgray}MemVP (7B)&9e-3&32&20&scaling  $\lambda = 0.01$, length of position embedding $n = 320$\\
\rowcolor{lightgray}MemVP (13B)&9e-3&32&20&scaling  $\lambda = 0.01$, length of position embedding $n = 400$\\
\bottomrule
\end{tabular}  
}

\label{tab:hyper-llama}  
\end{table} 


\section{Supplementary Experiments}

\subsection{Video Question Answering}
We conduct experiments on video question answering tasks on BART-base, including TVQA~\cite{tvqa} and How2QA~\cite{how2qa}. We follow the setting used for image-based tasks and fine-tune each dataset under single-task protocol. We use the features of \texttt{[CLS]} tokens extract by CLIP-ViT-B/32 from the 64 frames of each video. We train each method for 20 epochs with batch size 64. We report results on validation sets since the test sets are not publicly available. The results shown in Table~\ref{tab:video} verify the effectiveness of MemVP on video-based tasks.

\begin{table*}[h]  
\caption{{\bf Results on TVQA and How2QA.} ``FLOPs'' denotes the average FLOPs of language models on validation set.}  
\centering  
\resizebox{0.6\linewidth}{!}{
\begin{tabular}{lc|cc|cc}  
\toprule
\multirow{2}{*}{Method}& \multirow{2}{*}{\makecell[c]{\#Trainable \\Params (M/task)} }&\multicolumn{2}{c|}{TVQA} & \multicolumn{2}{c}{How2QA}   \\ 
&&Acc &FLOPs (G)&Acc&FLOPs (G)\\
\midrule
LoRA&2.74&68.2&44.6&69.4&38.1
\\
VL-Adapter&2.69&68.9&45.3&68.6&38.8\\
VL-PET&2.66&69.1&46.3&69.1&39.7\\
\rowcolor{lightgray}
MemVP (Ours)&\bf2.60&\bf69.7&\bf38.9&\bf70.7&\bf32.3\\
\bottomrule
\end{tabular}  
}

\label{tab:video}  
\end{table*} 
\subsection{Visual Instruction Tuning}
We also conduct experiments on visual instruction tuning of large language models. Following LLaVA-1.5~\cite{llava1.5}, we perform instruction tuning using MemVP on CLIP-ViT-L/14@336 and Vicuna-v1.5-7B with a two-stage protocol. In the first stage, we pretrain the projector and position embedding on LCS-558K for 1 epoch. In the second multi-task finetuning stage, we find that when the number of updated parameters is fewer than 100M, the model's instruction-following performance is notably poor. Therefore, we also insert LoRA into the language model, and jointly fine-tune the projector, position embedding, and LoRA modules with rank 128 on the 665K instruction tuning data used by LLaVA-1.5. In Table~\ref{tab:vit}, we report the performance of MemVP on various benchmarks, as well as the pretraining speed (batch size = 32/GPU) and inference speed on MMB dataset (batch size = 64/GPU).

\begin{table*}[t]  
\caption{{\bf Results of visual instruction tuning.}}  
\centering  
\resizebox{0.99\linewidth}{!}{
\begin{tabular}{l|ccc|ccccc|cc}  
\toprule
Method&Vision Encoder&Language Model&\# Trainable&GQA&SQA&MME-P&MMB&POPE&Pre-Training Time&Inference Time   \\ 
\midrule
LLaVA-1.5-7B&CLIP-ViT-L@336&Vicuna-v1.5-7B&7B&62.0&66.8&1511&64.3&85.9&5.43 s/batch&5.28 s/batch\\
MobileVLM-1.7B&CLIP-ViT-L@336&MobileLLaMA-1.4B-chat&1.4B&56.1&54.7&1196&53.2&84.5&0.61 s/batch&1.73 s/batch\\
MemVP-7B&CLIP-ViT-L@336&Vicuna-v1.5-7B&346M&58.7&69.0&1384&62.6&85.3&0.60 s/batch&1.71 s/batch\\
\bottomrule
\end{tabular}  
}

\label{tab:vit}  
\end{table*}

As the results in the above table, compared to LLaVA-1.5, MemVP is much more faster in training and inference. Although MemVP cannot compete with LLaVA-1.5 on some tasks, which highlights a limitation of our model, it is noteworthy that the pretraining and inference speed of MemVP-7B is comparable to MobileVLM-1.7B~\cite{mobilevlm} which is a smaller LLaVA-style model trained on the same data, but MemVP-7B outperforms MobileVLM-1.7B by a large margin.

\end{document}